%% file: ms.tex
\documentclass[letterpaper, 10 pt, conference]{ieeeconf} 
\usepackage{times} 

\usepackage{xspace} 
\makeatletter
\DeclareRobustCommand\onedot{\futurelet\@let@token\@onedot} 
\def\@onedot{\ifx\@let@token.\else.\null\fi\xspace} 
 
\def\wrt{w.r.t\onedot}  

\let\NAT@parse\undefined
\makeatother
\usepackage[numbers]{natbib}
\usepackage{algorithm}
\usepackage[algo2e,lined]{algorithm2e}
\usepackage{pseudocode}
\usepackage{multicol}
\usepackage[bookmarks=true]{hyperref}
\usepackage{color}
\usepackage{colortbl}
\usepackage{graphicx} % for pdf, bitmapped graphics files
\usepackage{amsmath} % assumes amsmath package installed
\usepackage{amssymb}  % assumes amsmath package installed
\usepackage{xcolor}
\usepackage{mdframed}
\usepackage{subfigure}
\usepackage{epstopdf}
\usepackage{siunitx}
\usepackage{booktabs}
\usepackage{multirow}
\usepackage[font={small}]{caption}
\usepackage{standalone}
\usepackage{epstopdf}
\usepackage{verbatim}
\usepackage{booktabs}
\usepackage{bm}

\SetKwInOut{Parameter}{Parameters}

\newcounter{parms}

\newcommand{\reffig}[1]{Fig.~\ref{#1}}
\newcommand{\refeq}[1]{Eq.~(\ref{#1})}
\newcommand{\reftab}[1]{Table~\ref{#1}}
\newcommand{\refsec}[1]{Section~\ref{#1}}
\newcommand{\refalg}[1]{Alg.~\ref{#1}}

\definecolor{todo-red}{RGB}{200,12,12}
\definecolor{green4}{RGB}{0,128,0}

\newcommand{\igor}[1]{\textcolor{orange}{[\textbf{igor:} #1]} }

\begin{document}

\title{\LARGE \bf
Visual-inertial Self-calibration on Informative Motion Segments
}
\author{\authorblockN{
Thomas Schneider\authorrefmark{1}, Mingyang Li\authorrefmark{2}, Michael Burri\authorrefmark{1}, Juan Nieto\authorrefmark{1}, Roland Siegwart\authorrefmark{1} and Igor Gilitschenski\authorrefmark{1}}
\authorblockA{\authorrefmark{1}Autonomous Systems Lab, ETH Zurich, \authorrefmark{2}Google Inc., Mountain View, CA}}

%------------------------------------------------------------
\maketitle
\thispagestyle{empty}
\pagestyle{empty}
%------------------------------------------------------------ 

\begin{abstract}
Environmental conditions and external effects, such as shocks, have a significant impact on the calibration parameters of visual-inertial sensor systems.
Thus long-term operation of these systems cannot fully rely on factory calibration.
Since the observability of certain parameters is highly dependent on the motion of the device, using short data segments at device initialization may yield poor results.
When such systems are additionally subject to energy constraints, it is also infeasible to use full-batch approaches on a big dataset and careful selection of the data is of high importance.

In this paper, we present a novel approach for resource efficient self-calibration of visual-inertial sensor systems. 
This is achieved by casting the calibration as a segment-based optimization problem that can be run on a small subset of informative segments.
Consequently, the computational burden is limited as only a predefined number of segments is used.
We also propose an efficient information-theoretic selection to identify such informative motion segments.
In evaluations on a challenging dataset, we show our approach to significantly outperform state-of-the-art in terms of computational burden while maintaining a comparable accuracy.

\begin{comment}
	\begin{itemize}
		\item Enrivonmental conditions, particularly the temperature (+vibrations, shock, possibly aging of electronics)  are crucial for calibarion
		\item Visual-inertial odometry systems become more and more popular in mobile robotics and augmented reality applications.
		\item Accurate calibration necessary for accurate state esitmation.
		\item Calibration parameters prone to environmental conditions and may change over the lifetime of a device.
		\item Thus, regular sensor calibration is required to maintain accurate state estimation
		\item Additionally, the underlying devices are typically resource constrained, including limited computational capabilities and limited energy supply.
		\item This paper presents a calibration approach that improves the estimates of the calibration parameters as new data becomes available during operation.
		\item In order to reduce computational burden, we propose a method selecting informative motion segments that are particularly suitable for calibration. Yielding similar results to full-batch based approaches at a considerably lower computational cost.
		\item We also show that our proposed segment selection methodology outperforms existing segment-based calibration approaches in terms of computational demand and energy consumption.
	\end{itemize}
\end{comment}
\end{abstract}

\section{Introduction}
\input{sections/0_introduction}

\section{Related Work}
\input{sections/1_related_work}

\section{Visual-inertial Models and Calibration}
\label{sec:batch_calib}
\input{sections/2_vi_calibration}

\section{Method}
\label{sec:method}
\input{sections/3_method}

\section{Experiments and Results}
\input{sections/4_results}

\section{Conclusions}
\input{sections/5_conclusions}

\section{Acknowledgement}
\input{sections/6_acknowledgement}
%------------------------------------------------------------
\bibliographystyle{ieeetr} % use IEEEtran.bst style
\small
\bibliography{robotvision}

% This command serves to balance the column
% lengths
\addtolength{\textheight}{-1cm}
\end{document}

%% file: sections/0_introduction.tex
%
% Motivation
%
In this work, we address the problem of sensor self-calibration of a visual-inertial tracking system, i.e., a state estimation system that fuses measurements from an inertial measurement unit (IMU) and one/multiple cameras to compute pose (position and orientation) estimates of a moving platform.
In recent years visual-inertial tracking has witnessed an ever increasing gain in popularity and is used in numerous mobile devices, virtual and augmented reality systems, and robotic platforms.
This success story results in large-scale projects such as Google Tango or Microsoft's HoloLens promising to make these complex systems available as part of consumer devices with a limited energy supply and may be operated by inexperienced users over a potential lifespan of several years.
These developments pose novel technical challenges to ensure accurate calibration of extrinsics and intrinsics of the underlying sensor systems.

%
% Calibration when there is no Janosch and no ASL around.
%
Outside a lab environment, varying environmental conditions (such as temperature) and a long lifespan result in changing calibration parameters that make permanent use of factory calibration infeasible even when assuming all parameters to be constant over a short or medium timespan.
In the absence of experienced engineers with access to special calibration routines and calibration patterns, the systems need to be capable of calibrating automatically in a potentially unknown environment.
Even though it was shown that calibration is also possible by using natural visual landmarks only~\cite{li2014high}, parameters such as axis misalignment of the IMU can only be observed under certain motion.
One possible solution is to run a full-batch calibration procedure over as much data as possible.
However, this results in a huge computational load making this methodology infeasible for consumer devices with limited computational resources and a limited power supply.
\begin{figure}
	\centering
	\includegraphics[width=0.5\textwidth]{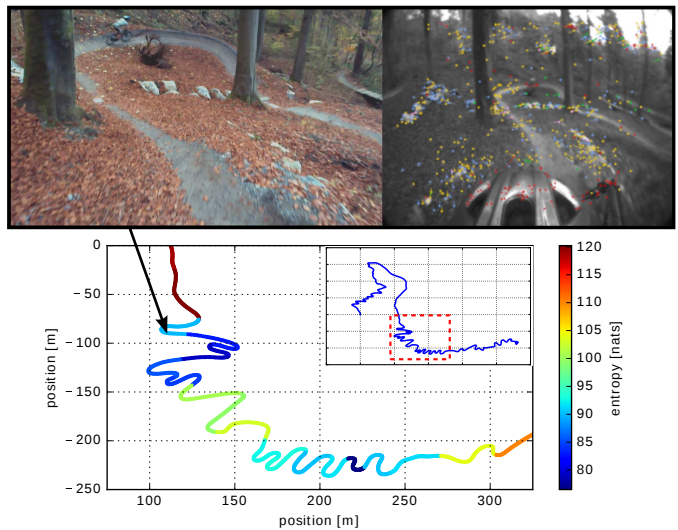}
	\caption{Riding down Mount Uetliberg on a mountain-bike with a camera and IMU attached to the rider's helmet:
		This dataset is a good illustration of the vastly varying amount of information available in different segments of the trajectory.
		The color indicates the information content of the segment \wrt the sensor calibration parameters (intrinsics and extrinsics of camera/IMU) where a lower value indicates more information.
		Consequently, the information measure is used to sparsify the sensor self-calibration problem by excluding less informative portions of the dataset.
	}
	\label{fig:teaser}
\end{figure}

%
% General Idea of proposed approach
%
This work makes use of the fact that visual-inertial estimation systems typically run for a sufficiently long time to perform a lot of different types of motion eventually.
Therefore, our system is designed to automatically select informative motion segments, that are well suited for calibration.
The information measure, used for this identification, is illustrated on an example trajectory in \reffig{fig:teaser}.
The most informative segments are then stored in a database and used to refine the calibration from time to time.
This not only helps in getting good calibration data, but also reduces the size of the calibration problem considerably.
Furthermore, we show that the results of our calibration, using only a small number of segments, is comparable in accuracy to the results obtained with a full batch approach over all the data collected.

%
% Contributions
%
This paper makes the following contributions:
\begin{itemize}
	\item We present an efficient information-theoretic procedure to identify the most informative segments of a trajectory.
	\item We propose a segment-based method for self-calibration of the intrinsic and extrinsic parameters of visual-inertial sensor systems.
	\item In thorough evaluations we show that the proposed methodology achieves comparable results to a full batch approach and state-of-the-art while at the same time requires a significant lower complexity and computational effort.
\end{itemize}

%% file: sections/1_related_work.tex
%%
% Motivation: VIO/SLAM
Over the last decade, visual-inertial SLAM has received great attention from the research community and tremendous progress has been achieved.
For example, the work of \cite{leutenegger2015keyframe} demonstrates a fixed-lag-smoother based VIO framework, that achieves accuracies in the sub-percent range over the travelled distance.
However, on constrained platforms such as mobile phones, filtering based algorithms are preferred such as \cite{li2013high} and \cite{bloesch2015robust} that show similar accuracies at lower computational complexity.

%%
% Traditional calibration approaches and models.
To achieve such accuracies, precise calibration of the sensor models is required.
Traditionally, camera models are calibrated using a calibration target such as in the work of \cite{zhang2000flexible}.
It has also been shown that camera models can be obtained using natural features only \cite{devernay2001straight}.
The increasing usage of low-cost MEMS IMUs further requires calibration of the inertial sensors, referred to as IMU intrinsics.
The work of \cite{krebs} presents an inertial model, which we will adopt in this work, that considers scale inaccuracies and misalignments of individual sensors axes.
In \cite{nikolic2016non} a batch estimator is presented that calibrates the latter model relying on a calibration pattern.
The model of \cite{rehder2016extending} additionally considers the location of individual accelerometer axes where the parameters are estimated in a continuous-time formulation using a parametric estimation framework.

%%
% Self-calibration and calibration performed by non-expert users.
The recent roll-out of advanced SLAM systems to a wide audience creates a need for simple calibration algorithms accessible to users without access to special equipment such as calibration targets.
The work of \cite{li2014high} mitigates these short-comings by including the calibration parameters directly into an EKF-based VIO estimator and performing visual and inertial self-calibration solely based on natural features.
%%
% Non-expert users - observability aware calibration.
Visual inertial systems, however, require special motion in order to render all calibration parameters observable \cite{mirzaei2008kalman}.
Therefore, observability-aware calibration methods have been developed to aid non-expert users in collecting a complete dataset of minimal size and improve the estimation quality.
In \cite{maye2013self}, a set of informative segments is selected using an information-gain measure to consequently perform a calibration over this set.
Further, a truncated QR solver is used to constrain parameter updates to the observable sub-space.
The generality of this method makes it applicable to a wide-range of estimation problems.
Unfortunately, the evaluation of the utilized information metric is expensive and can prevent its use especially on resource constrained platforms.
In our work, we follow a similar approach and identify informative motion segments to build a sparser but complete calibration dataset.
Similarly to the work of \cite{keivan2014constant}, we use the entropy to efficiently approximate the information content of segments but calibrate the full visual-inertial model instead of just a camera.
Additionally, we extend the information measure and evaluate the informativeness of segments \wrt to subgroups of the high-dimensional parameter vector and thus mitigate the drawback of using a scalar measure.
Another interesting approach approximates the information of a trajectory segment by the local observability Gramian, as described in \cite{hausman2016observability}, where it is used in an active calibration setting.

%% file: sections/2_vi_calibration.tex
In this section, we will introduce the sensor models for the camera and IMU and formulate the batch estimation problem for self-calibration.~\footnote{It is important to note that the method described in this paper generalizes to arbitrary problems, however it is presented on the application of visual-inertial self-calibration.}

\subsection{Notation and Frames of Reference}
A transformation matrix $\mathbf{T}_{AB}\in\mathbb{SE}^3$ takes vector $_{B}{\mathbf{p}}\in\mathbb{R}^3$ from the frame of reference $B$ to the frame of reference $A$ and can be further partitioned into a rotation matrix $\mathbf{R}_{AB}\in\mathbb{SO}^3$ and a translation vector $_{A}{\mathbf{p}}_{AB}\in\mathbb{R}^3$ as follows:
\begin{equation}
\label{eq:trafo}
\begin{bmatrix}
_{A}{\mathbf{p}} \\ 1
\end{bmatrix}
 = \mathbf{T}_{AB} \cdot 
\begin{bmatrix}
_{B}{\mathbf{p}} \\ 1
\end{bmatrix}
= 
\begin{bmatrix}
\mathbf{R}_{AB} & _{A}{\mathbf{p}}_{AB}\\ 
 \mathbf{0} & 1 
\end{bmatrix}
\cdot 
\begin{bmatrix}
_{B}{\mathbf{p}} \\ 1
\end{bmatrix}
\end{equation}
Further, the unit quaternion $\mathbf{q}_{AB}$ represents the rotation corresponding to $\mathbf{R}_{AB}$ as defined in \cite{trawny2005indirect}.
The operator $\mathbf{T}_{AB}(\cdot)$ is defined to transform a vector in $\mathbb{R}^3$ from $B$ to the frame of reference $A$ as $_{A}{\mathbf{p}} = \mathbf{T}_{AB} \left( _{B}{\mathbf{p}} \right)$ according to \refeq{eq:trafo}.

\begin{figure}[]
\centering
\includegraphics[width=0.35\textwidth]{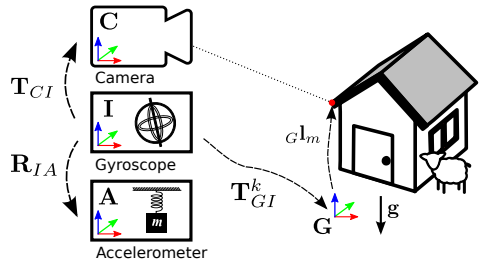}
\caption{Frame of reference definitions for the visual-inertial system. A camera, 3-dof accelerometer and 3-dof gyroscope are rigidly attached to an agent. The estimated pose of the agent at timestep $k$ is expressed by the transformation $\mathbf{T}_{GI}^k$. A 6-dof transformation matrix $\mathbf{T}_{CI}$ relates the gyroscope's frame $I$ to the camera's frame $C$. The accelerometer frame $A$ is only rotated \wrt gyroscope's frame $I$ by $\mathbf{R}_{IA}$, since $_{I}{\mathbf{p}}_{IA}$ in single-chip MEMS-IMUs is typically close to zero.}
\label{fig:frames_of_ref}
\end{figure}

The \reffig{fig:frames_of_ref} illustrates the relevant coordinate frames used within this work.
The frame $G$ denotes a gravity aligned ($_G{\mathbf{e}}_z=-\mathbf{g}$) inertial frame and is used to express the estimated pose of the agent $\mathbf{T}_{GI}^k$ and the position of the estimated landmarks $_G{\mathbf{l}_m}$.
The frame $I$ coincides with the sensing axes of the gyroscope and is chosen as the body frame of the agent.
The camera frame $C$ and accelerometer frame $A$ are rigidly attached to the body frame.
The extrinsic calibration transformations for the camera $\mathbf{T}_{CI}$ and the rotation matrix for the accelerometer $\mathbf{R}_{IA}$ are to be estimated and are both defined relative to the frame of the gyroscope $I$ that is used as the body frame. 

%%%%%%%%%%%%%%%%%%%%%%%%%%%%%%%%%%%%%%%%%%%%%%%%%%%%%%%%%%%%%%%%%%%%%%%%%%%%%%%%%%%%%%%%%%%%%%%%%%%%%%%%%%%%%%%%
\subsection{Inertial Model}
\label{sec:model_inertial}
A triad of (ideally) orthogonal gyroscopes are used to sense the true angular velocities $_I\bm{\omega}_{GI}$ of the body frame $I$ \wrt the world-fixed inertial frame $G$.
The gyroscope measurements $\tilde{\bm{\omega}}$ are modeled similar to \cite{krebs,li2014high} as:
\begin{equation}
\tilde{\bm{\omega}} = \mathbf{T}_g \cdot _{I}\bm{\omega}_{GI} + \mathbf{b}_g + \bm{\eta}_g
\end{equation}
where the bias $\mathbf{b}_g$ follows a random walk process as $\dot{\mathbf{b}}_g = \bm{\eta}_{bg}$ and $\bm{\eta}_{g}$ and $\bm{\eta}_{bg}$ are zero-mean, white Gaussian noise processes.
The matrix $\mathbf{T}_g$ accounts for scale errors and sensor axis misalignments present in cheaper sensors. It is assumed to be a constant over time and is structured as:
\begin{equation}
  \label{eq:gyrocalib}
  \mathbf{T}_g = \begin{bmatrix}
  s^x_g & m^x_g & m^y_g \\ 
  0 & s^y_g  & m^z_g \\ 
  0 & 0 & s^z_g
  \end{bmatrix},
  \mathbf{s}_g = \begin{bmatrix}
  s^x_g \\ s^y_g \\ s^z_g
  \end{bmatrix},
  \mathbf{m}_g = \begin{bmatrix}
  m^x_g \\ m^y_g \\ m^z_g
  \end{bmatrix}
\end{equation}
with $\mathbf{s}_g$ and $\mathbf{s}_g$ denoting the collection of all parameters from $\mathbf{T}_g$.

Similarly, the specific force measurements $\tilde{\mathbf{a}}$ of the accelerometer are modeled as:
\begin{equation}
\tilde{\mathbf{a}} = \mathbf{T}_a \cdot \mathbf{R}_{AI} \cdot \mathbf{R}_{IG}^k \cdot \left( _{G}{\mathbf{a}}_{GI} - _{G}{\mathbf{g}} \right) + \mathbf{b}_a + \bm{\eta}_a
\end{equation}
where $\mathbf{T}_a$ is a calibration matrix and $\mathbf{b}_a$ defines a random walk process analog to the gyroscope model.
The calibration states for the IMU models can be summarized as:
\begin{equation}
  \bm{\theta}_i = \begin{bmatrix}
  \mathbf{s}_g^T & \mathbf{m}_g^T & \mathbf{s}_a^T & \mathbf{m}_a^T & \mathbf{q}_{AI}^T
  \end{bmatrix}^T
\end{equation}

It is important to note that the values of the scale parameters $s_i$ can't be used directly to correct the scales of each individual axis, instead a linear combination of all factors applies.
Further details can be found in \cite{li2014high}.

%%%%%%%%%%%%%%%%%%%%%%%%%%%%%%%%%%%%%%%%%%%%%%%%%%%%%%%%%%%%%%%%%%%%%%%%%%%%%%%%%%%%%%%%%%%%%%%%%%%%%%%%%%%%%%%%
\subsection{Camera Model} 
\label{sec:model_cam}
Let $_{G}\mathbf{l}_m$ denote a 3-d landmark observed from keyframe $k$ that is projected into a 2-d point $\mathbf{z}_{k,m}$ on the image plane of the camera as follows:
\begin{equation}
\begin{aligned}
\mathbf{z}_{k,m}(\mathbf{T}_{IG}, \mathbf{l}_m, \bm{\theta}_c) = f_p(\bm{\theta}_c, \mathbf{T}_{CI}(\mathbf{T}_{IG}( _{G}\mathbf{l}_m))) + \bm{\eta}_{c}
\end{aligned}
\end{equation}
where $f_p(\cdot)$ denotes the perspective projection function and $\bm{\eta}_c \sim \mathcal{N}(\mathbf 0, \sigma_{c}^2 \cdot \mathbf{I}_{2})$ a white Gaussian noise process.

For the evaluations, we parametrize the projection function $f_p$ using a pinhole camera model and field-of-view (FOV) distortion model of \cite{devernay2001straight}.
The calibration state relevant for the camera model then is:
\begin{equation*}
  \bm{\theta}_c = \begin{bmatrix}
  {\mathbf{q}_{CI}}^T & {_{C}{\mathbf{p}}_{CI}}^T & \mathbf{f}^T & \mathbf{c}^T & w
  \end{bmatrix}^T
\end{equation*}
where $\mathbf{q}_{CI}$ and $_{C}{\mathbf{p}}_{CI}$ are the extrinsic calibration of the camera \wrt IMU, $\mathbf{f}=\begin{bmatrix}f_x & f_y\end{bmatrix}^T$ the focal lengths, $\mathbf{c}=\begin{bmatrix}c_x & c_y\end{bmatrix}^T$ the principal point and $w$ a distortion parameter.

% FOV distortion function.
%\begin{equation}
%\mathbf{f_p\left(\mathbf{l}\right)} = 
%\begin{bmatrix}
%\tfrac{f_x}{w} \cdot \tfrac{l_x}{\sqrt{{l_x}^2 + {l_y}^2}} \cdot atan \left( %\tfrac{\sqrt{{l_x}^2 + {l_y}^2}}{l_z} \cdot 2 \cdot tan(\tfrac{w}{2}) \right ) + c_x
%\\ 
%\tfrac{f_y}{w} \cdot \tfrac{l_y}{\sqrt{{l_x}^2 + {l_y}^2}} \cdot atan \left( %\tfrac{\sqrt{{l_x}^2 + {l_y}^2}}{l_z} \cdot 2 \cdot tan(\tfrac{w}{2}) \right ) + c_y
%\end{bmatrix}
%\end{equation}

%%%%%%%%%%%%%%%%%%%%%%%%%%%%%%%%%%%%%%%%%%%%%%%%%%%%%%%%%%%%%%%%%%%%%%%%%%%%%%%%%%%%%%%%%%%%%%%%%%%%%%%%%%%%%%
\subsection{Maximum-likelihood Estimator}
\label{sec:ml_calib}
The framework of maximum-likelihood estimation is used to jointly estimate all keyframe states $\mathbf{x}_k$ (\refeq{eq:keyframe_states}), the scene as a set of observed point landmarks ${_G}{\mathbf{l}_m}$, the calibration parameters of the camera $\bm{\theta}_c$ and the IMU $\bm{\theta}_i$ with the keyframe state $\mathbf{x}_{k}$ being defined as: 
\begin{equation}
\label{eq:keyframe_states}
\mathbf{x}_{k}=\begin{bmatrix} {\mathbf{q}_{GI}^{k}}^T & {_{G}\mathbf{p}_{GI}^{k}}^T & {_{G}\mathbf{v}_{I}^{k}}^T & {\mathbf{b}_{a}^{k}}^T & {\mathbf{b}_{g}^{k}}^T \end{bmatrix}^T
\end{equation}
where $\mathbf{q}_{GI}^{k}$ and $_{G}\mathbf{p}_{GI}^{k}$ denote the pose of the agent, $_{G}\mathbf{v}_{I}^{k}$ the velocity of the IMU expressed in frame $G$, and $\mathbf{b}_{\cdot}^{k}$ the biases for the gyroscope or accelerometer. 
For convenience of notation, the individual states are stacked into vectors as follows:
\begin{equation*}
\begin{aligned}
&\mathbf{\hat{X}} =
\begin{bmatrix}
{\mathbf{\hat x}_0}^T & \hdots & {\mathbf{\hat x}_K}^T
\end{bmatrix}^T, 
\quad
\mathbf{\hat{L}} =
\begin{bmatrix}
{_G\mathbf{\hat l}_M}^T & \hdots & {_G\mathbf{\hat l}_M}^T
\end{bmatrix}^T, \\
&\bm{\hat{\theta}} = \begin{bmatrix} {\bm{\hat{\theta}}_c}^T & {\bm{\hat{\theta}}_i}^T \end{bmatrix}^T  
\end{aligned}
\end{equation*}
where $K$ denotes the number of keyframes and $M$ the number of landmarks.
Additionally $\bm{\hat{\pi}}$ defines the collection of all estimated quantities as:
\begin{equation*}
\begin{aligned}
\bm{\hat{\pi}} &= \begin{bmatrix} \bm{\hat{\theta}}^T & \mathbf{\hat{X}}^T & \mathbf{\hat{L}}^T \end{bmatrix}^T
\end{aligned}
\end{equation*}
We want to infer $\bm{\pi}$ from measurements $\mathbf{z}_{k,m}$ made by a camera and measurements $\mathbf{u}_k$ of an IMU. The stacked vector forms of the measurements  are defined as follows:
\begin{equation*}
\begin{aligned}
\mathbf{Z}&=\{\mathbf{z}_{k,m} | k \in [0, K], m \in [0, M(k)]\} \\
\mathbf{U}&=\{\mathbf{u}_k | k \in [0,K-1]\}
\end{aligned}
\end{equation*}

\begin{figure}
\centering
\includegraphics[width=0.28\textwidth]{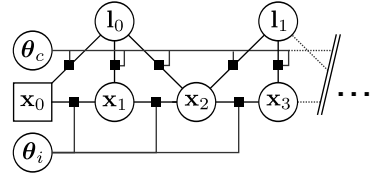}
\caption{%
Calibration problem in factor graph representation that contains visual-inertial keyframe states $\mathbf{x}_k$ (pose, velocity and IMU biases), landmarks $\mathbf{l}_m$ and calibration states for the camera $\bm{\theta}_c$ and IMU $\bm{\theta}_i$. 
The square around the initial node $\mathbf{x}_0$ denotes a gauge fix of the position $_{G}{\mathbf{p}}_{GI}$ and rotation around the gravity vector.
Integrated IMU measurements constitute the inertial factor $g^{imu}_{k}(\mathbf{x}_k, \mathbf{x}_{k-1}, \bm{\theta}_i, \mathbf{u}_k)$ and the landmark projection factors $g^{cam}_{k,m}(\mathbf{x}_k, \mathbf{l}_m, \bm{\theta}_c, \mathbf{z}_{k,m})$ models the camera measurements.
}
\label{fig:factor_graph}
\end{figure}
Following the sensor models described in \refsec{sec:model_inertial}-\ref{sec:model_cam}, a probability model is defined as shown in \reffig{fig:factor_graph}.
Probabilistic inertial constraints $g^{imu}_{k}$ between consecutive keyframe states $k$ and $k+1$ are formed as a function of the integrated IMU measurements and the corresponding measurement uncertainties \cite{li2013high}.
The likelihood $p(\cdot)$ of this model can be expressed as:
\begin{equation}
\begin{split}
\label{eq:prob_model}
p(\bm{\pi} | \mathbf{Z},\mathbf{U}) \varpropto &\prod_{k=1}^{K} p(\mathbf{x}_k|\mathbf{x}_{k-1},\bm{\theta}_i,\mathbf{u}_k) \\
\cdot & \prod_{k=0}^{K} \prod_{m=0}^{M(k)} p(\mathbf{z}_{k,m}|\mathbf{x}_k,\mathbf{l}_{m}, \bm{\theta}_c) \\
\end{split}
\end{equation}
where $p(\mathbf{x}_k|\mathbf{x}_{k-1},\bm{\theta}_i,\mathbf{u}_i)$ denotes the inertial constraints between two consecutive keyframe states as a function of integrated IMU measurements $\mathbf{u}_k$ and $p(\mathbf{z}_{k,m}|\mathbf{x}_k,\mathbf{l}_{m})$ the measurement model of the point landmark observation $\mathbf{z}_{k,m}$ of the $m$-th landmark observed from the $k$-th keyframe. 
More details on the derivation can be found in \cite{nikolic2016non}.

The maximum-likelihood (ML) estimate $\bm{\hat{\pi}}_{ML}$ is obtained by solving the optimization problem that maximizes the likelihood of \refeq{eq:prob_model}:
\begin{equation}
\label{eq:optim}
\bm{\hat{\pi}}_{ML} = \displaystyle{\underset{\bm{\pi}}{\operatorname{argmax}}\,p(\bm{\pi}|\mathbf{Z},\mathbf{U})}
\end{equation}
With the assumptions of Gaussian noise for all sensor models, as discussed in \refsec{sec:model_inertial} and \refsec{sec:model_cam}, the optimization problem defined in \refeq{eq:optim} is equivalent to a non-linear least squares problem.
This problem can be solved using numerical minimization approaches, where standard methods include Gauss-Newton, Levenberg-Marquardt, Dogleg, etc.
In our implementation, we use the Levenberg-Marquardt implementation of the Ceres framework \cite{ceres_solver}.

%In this case, the iterative update rule, starting from an initial linearization point $\mathbf{\hat{X}}_0$, can be written as:
%\begin{equation}
%\label{eq:nls}
%\begin{aligned}
%\left( \mathbf{J}^{T} \mathbf{T}^{-1} \mathbf{J}\right) \delta \bm{\pi}_k &= \mathbf{J}^{T} %\mathbf{T}^{-1} \mathbf{e(\hat{\bm{\pi}}_k)} \\
%\bm{\hat{\pi}}_{k+1} &= \bm{\hat{\pi}}_{k} \boxplus \delta \bm{\hat{\pi}}_k 
%\end{aligned}
%\end{equation}
%where $\mathbf{T}$ is the error covariance matrix that contains the motion and sensor model covariance blocks,
%$\mathbf{J}$ stacks the Jacobians of all error terms. The operator $\boxplus$ denotes the addition of a small delta state $\mathbf{\delta \hat{X}_{k}}$ in minimal coordinates which is equivalent to a normal addition for all quantities except quaternions where a small angle approximation is used as defined in \cite{trawny2005indirect}.
%
%The expression $\mathbf{J}^{T} \mathbf{T}^{-1} \mathbf{J}$ evaluated with $\mathbf{\hat{X}}$ at convergence is called the Fisher Information Matrix and corresponds to the inverse of the covariance of $\mathbf{\hat{X}}$.
%
%In our implementation, we use the Levenberg-Marquardt algorithm because of its robustness and regularizing property \cite{marquardt1963algorithm}.

%% file: sections/3_method.tex
\begin{figure}[]
\centering
\includegraphics[width=0.42\textwidth]{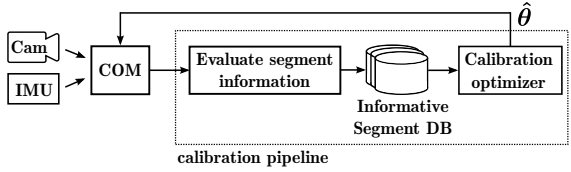}
\caption{System overview and context: informative motion segments are identified from the output of an existing ego-motion estimator (COM) and maintained in a database for future calibration.}
\label{fig:arch}
\end{figure}

%
% Context where the method runs.
%
The proposed self-calibration method aims at being run in parallel to an existing visual-inertial SLAM system that provides motion estimates as shown in \reffig{fig:arch}.
In our implementation we use a concurrent odometry and mapping (COM) framework consisting of \cite{li2013high}, \cite{nerurkar2014c} and \cite{lynen2015get} but it is important to note that the proposed algorithms are not tied to a particular SLAM formulation.
The SLAM system uses a calibration from previous runs or nominal values for the device at hand.~\footnote{If no priors are available, a complete self-calibration may be difficult and specialized initialization techniques should be used beforehand e.g.~\cite{zhang2000flexible,hughes2010equidistant}.}
%
% Algorithm
%
The stream of estimated keyframes $\mathbf{\hat x}_i$ and landmarks $\mathbf{\hat l}_i$ leaving the COM module is partitioned into motion segments $\mathcal{S}^i$ of a predefined size $N$ as follows:
\begin{equation}
\begin{split}
\mathbf{\hat X}_{\mathcal{S}}^i &=
  \begin{bmatrix}
  \mathbf{\hat x}_i^T, & \cdots, & \mathbf{\hat x}_{i+(N-1)}^T
  \end{bmatrix}^T \\
\mathbf{\hat L}_{\mathcal{S}}^i &=
  \begin{bmatrix}
  \mathbf{\hat l}_i^T, & \cdots, & \mathbf{\hat l}_{i+(N-1)}^T
  \end{bmatrix}^T
\end{split}
\end{equation}
where $\mathbf{\hat X}_{\mathcal{S}}^i$ denotes the keyframes within the $i$-th segment and $\mathbf{\hat L}_{\mathcal{S}}^i$ the landmarks observed by the $i$-th segment.
An efficient information-theoretic measure is used to evaluate each new candidate segment for their information content \wrt the calibration parameters and the most informative segments are maintained in a database.
Once enough segments have been collected, an ML-based calibration is triggered to estimate the calibration parameters.
An overview of the algorithm is shown in \refalg{alg:calib_algo}.
The remainder of this section will discuss the algorithm in more detail.
\begin{algorithm}[]
 \definecolor{commentcolor}{RGB}{150,0,0}
 \footnotesize
 \KwIn{Initial calibration: $\bm{\hat\theta}_{init}$}
 \KwOut{Updated calibration: $\bm{\hat\theta}$}
 \SetKwFunction{waitsensor}{WaitForNewSensorData}
 \SetKwFunction{runcom}{RunCOM}
 \SetKwFunction{collect}{CollectNKeyframes}
 \SetKwFunction{evalinfo}{EvaluateSegmentInformation}
 \SetKwFunction{adddb}{UpdateDatabase}
 \SetKwFunction{dbfull}{EnoughSegmentsInDatabase}
 \SetKwFunction{runcalib}{RunOptimization}
 \SetKwFunction{getdb}{GetAllSegmentsFromDatabase}
 \SetKwFor{Loop}{Loop}{}{EndLoop}

 \BlankLine
 \Loop{} { 
  \textcolor{commentcolor}{// Initialize motion segments of size N from COM output.}\\
  $\mathcal{S}_{i} \leftarrow \{\}$  \\
  \Repeat{$dim(\mathcal{S}_{i}) == N$}{
    $data$ = \waitsensor{} \\
    $\mathbf{\hat x}_j$, $\mathbf{\hat l}_j \leftarrow$ \runcom{$data$, $\hat{\theta}_{init}$} \\
    $\mathcal{S}_{i} \leftarrow \mathcal{S}_{i} \cup (\mathbf{\hat x}_j$, $\mathbf{\hat l}_j)$
  }
  \BlankLine
  $H\left(\bm{\theta}\right) \leftarrow$ \evalinfo{$\mathcal{S}_i$} \textcolor{commentcolor}{// \refsec{sec:eval_entropy}}\\ 
  \adddb{$\mathcal{S}_{i}$, $H\left(\bm{\theta}\right)$} \textcolor{commentcolor}{// \refsec{sec:info_db}}\\    
  \If{\dbfull{}} {
    $\mathcal{\mathbf{S}}_{info} \leftarrow$ \getdb{} \\
    $\bm{\hat \theta} \leftarrow$ \runcalib{$\mathcal{\mathbf{S}}_{info}$}   \textcolor{commentcolor}{// \refsec{sec:sparified_problem}} \\
	\KwRet $\bm{\hat \theta}$
  }
  $i \leftarrow i + 1$
 }
 
 \caption{\small Method shown for a single parameter group}
 \label{alg:calib_algo}
\end{algorithm}

%%%%%%%%%%%%%%%%%%%%%%%%%%%%%%%%%%%%%%%%%%%%%%%%%%%%%%%%%%%%%%%%%%%%%%%%%%%%%%%%%%%%%%%%%%%%%%%%%%%%%%%%%%%%%%%%%%%%%%%%%%%%%%%%%
\subsection{Evaluating Information Content of Segments}
\label{sec:eval_entropy}
We use the differential entropy to quantify the information content of the $i$-the candidate segment $\mathcal{S}_i$ \wrt the calibration parameters $\bm{\theta}$ by considering only the constraints within each segment.
Using the entropy to evaluate the information of a candidate segments, as a score that is independent of all other segments, makes its evaluation very efficient at the cost that information coming from other segments is neglected.
For example, loop-closure constraints cannot be considered in the score, however, loop-closures are considered during calibration.

% Computation of the marginal covariance.
To calculate the segment entropy, we first approximate the covariance matrix of all states in the segment by the inverse of the Fisher Information Matrix as: 
\begin{equation}
\label{eq:segment_cov_fim}
\bm{\Sigma}_{\mathbf{XL}\bm{\theta}} = \operatorname{Cov}\left[ p(\mathbf{X}_{\mathcal{S}}^i, \mathbf{L}_{\mathcal{S}}^i, \bm{\theta} | \mathbf{U}_i,\mathbf{Z}_i) \right] = (\mathbf{J}_i^T \mathbf{T}_i^{-1} \mathbf{J}_i)^{-1} 
\end{equation}
where $\mathbf{J}_i$ denotes the Jacobian of all error-terms in the segment and $\mathbf{T}_i$ the stacked error-term covariance where the column ordering is chosen that the calibration parameters $\bm{\theta}$ lie on the right side.
To avoid a costly inversion of \refeq{eq:segment_cov_fim}, which becomes intractable for larger problems, we make use of a rank-revealing QR decomposition to obtain $\mathbf{Q_i}\mathbf{R_i}=\mathbf{L}_i\mathbf{J}_i$ where $\mathbf{T}_i^{-1}=\mathbf{L}_i^T\mathbf{L}_i$ denotes the Cholesky decomposition of the error-term covariance matrix.
\refeq{eq:segment_cov_fim} can then be rewritten as:
\begin{equation}
\label{eq:cholesky}
\bm{\Sigma}_{\mathbf{XL}\bm{\theta}} = (\mathbf{R}_i^T\mathbf{R}_i)^{-1} = 
\begin{bmatrix}
 \bm{\Sigma}_{\mathbf{XL}} & \bm{\Sigma}_{\mathbf{XL},\bm{\theta}} \\ 
\bm{\Sigma}_{\mathbf{XL},\bm{\theta}}^T & \bm{\Sigma}_{\bm{\theta}} \\ 
\end{bmatrix}
\end{equation}
In the context of sensor calibration, the keyframe $\mathbf{X}_{\mathcal{S}}^i$ and landmark states $\mathbf{L}_{\mathcal{S}}^i$ are considered nuisance variables and we are only interested in the marginal covariance $\bm{\Sigma}_{\bm{\theta}} = \operatorname{Cov} \left [p(\bm{\theta} | \mathbf{U}_i,\mathbf{Z}_i) \right]$ of the calibration parameters $\bm{\theta}$.
As $\mathbf{R}_i$ is an upper-triangular matrix, we can efficiently obtain the marginal covariance $\bm{\Sigma}_{\bm{\theta}}$ by back-substitution.

%Normalization
Before calculating the entropy, we normalize the marginal covariance $\bm{\Sigma}_{\bm{\theta}}$ to account for the different scales of the calibration parameters.
The normalized covariance $\bm{\bar \Sigma}_{\bm{\theta}}$ is calculated as:
\begin{equation}
\label{eq:cov_norm}
\bm{\bar \Sigma}_{\bm{\theta}} = \operatorname{diag}(\bm{\sigma}_{ref})^{-1} \cdot \bm{\Sigma}_{\bm{\theta}} \cdot \operatorname{diag}(\bm{\sigma}_{ref})^{-1}
\end{equation}
where $\bm{\sigma}_{ref}$ is the expected standard deviation of $\bm{\hat \theta}$ and was obtained from statistics over multiple reference segments. \
%Entropy.
The differential entropy $H\left(\bm{\theta}\right)$ of the normalized multivariate normal distribution $\bar p_{\bm{\theta}}(\bm{\theta})=\bar p(\bm{\theta} | \mathbf{U}_i,\mathbf{Z}_i)$ can then be calculated as:
\begin{equation}
\label{eq:entropy}
\begin{aligned}
H\left(\bm{\theta}\right) &= -\int _{-\infty }^{\infty } \cdots \int _{-\infty }^{\infty } \bar p_{\bm{\theta}}(\bm{\theta})\ln \bar p_{\bm{\theta}}(\bm{\theta})\,d\bm{\theta} \\
&={\frac {1}{2}}\ln \left((2\pi e)^{k}\cdot \operatorname{det}\left({\bm{\bar \Sigma}_{\bm{\theta}} }\right)\right),\\
\end{aligned}
\end{equation}
where $k$ is the dimension of the normal distribution.

\begin{figure}[]
\centering
\includegraphics[width=0.3\textwidth]{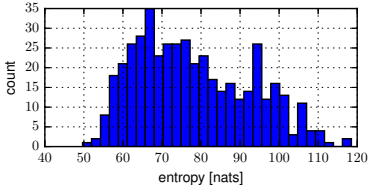}
\caption{Histogram of normalized segment entropies $H\left(\bm{\theta}\right)$ over $450$ segments from $15$ datasets.}
\label{fig:score_histogram}
\end{figure}

% Motivation sub-groups.
The segment entropy $H\left(\bm{\theta}\right)$ is not a directional measure and thus summarizes the information of all parameters $\bm{\theta}$ in a single scalar value.
For high-dimensional calibration vectors $\bm{\theta}$, however, the contribution of well-observable modes to the entropy might shadow weaker modes despite normalization.
This effect causes the distribution of the entropies to remain multimodal (as shown in \reffig{fig:score_histogram}) because the number of informative segments vs. less informative segments is in general not distributed equally within a given dataset.

\begin{comment}
	\begin{figure}[]
	\centering
	\includegraphics[width=0.4\textwidth]{plots/best-worst-segment.eps}
	\caption{Example trajectory of the most and least informative motion segment chosen from the histogram in \reffig{fig:score_histogram}.}
	\label{fig:traj_ex}
	\end{figure}
\end{comment}

%
% Sub-queues.
For this reason, the vector of calibration parameters $\bm{\theta}$ is partitioned into $Q$ sub-vectors $\bm{\theta}_q$ as:
\begin{equation}
\bm{\theta} = \begin{bmatrix} \tilde{\bm{\theta}}_{0}^T & \ldots & \tilde{\bm{\theta}}_{Q}^T \end{bmatrix}^T
\end{equation}
The marginal entropy is calculated for each parameter group $q$ using the corresponding marginal covariance $\bm{\bar \Sigma}_{\tilde{\bm{\theta}}_{q}}$ as described in \refeq{eq:entropy}.
The marginal segment entropies $H\left(\bm{\theta}_q\right)$ are then directly used as a measure of information contained in the segment \wrt to the parameters of group $q$ (where a lower entropy corresponds to richer information).

In this work, we partition the parameters $\bm{\theta}$ into three groups by sensor:
\begin{equation}
\begin{aligned}
\tilde{\bm{\theta}}_{0}  &= \begin{bmatrix}
  \mathbf{s}_g^T & \mathbf{m}_g^T & \mathbf{s}_a^T & \mathbf{m}_a^T & \mathbf{q}_{AI}^T
\end{bmatrix}^T \\
\tilde{\bm{\theta}}_{1} &=
\begin{bmatrix}
\mathbf{f}^T & \mathbf{c}^T & w
\end{bmatrix}^T  \\
\tilde{\bm{\theta}}_{2} &= 
\begin{bmatrix}
{\mathbf{q}_{CI}}^T & {_{C}{\mathbf{p}}_{CI}}^T 
\end{bmatrix}^T
\end{aligned}
\end{equation}
This follows the intuition that the problem exhibits different co-observability structures i.e. a set of parameters is always observable as a group e.g. the camera model requires only minimal motion (once the landmarks are initialized) whereas the inertial model requires sufficient excitation.
A more thorough analysis of how to identify the co-observability structure and thus optimally group the parameters should be part of future work.

%%%%%%%%%%%%%%%%%%%%%%%%%%%%%%%%%%%%%%%%%%%%%%%%%%%%%%%%%%%%%%%%%%%%%%%%%%%%%%%%%%%%%%%%%%%%
\subsection{Collecting Informative Segments in a DB}
\label{sec:info_db}
% Multiple DB with param groups
A database with $Q$ tables is maintained where each table retains the $N$ most informative segments for the corresponding parameter group $q$.
Segments can be in multiple tables if it is informative \wrt multiple parameter groups.
Therefore, the complexity of the calibration problem has an upper bound, as the max. number of segments in the database can be $Q\cdot N$ (or less if segments are in multiple tables).

% Discussion.
It is important to note that the sum of segment entropies is a conservative approximation to the true information in the database for two reasons:
First, the entropy is a scalar that ``summarizes`` the information of several parameters and thus does not contain any directional information.
Second, for efficiency, the segment entropy is calculated by neglecting the cross-terms to other segments.
This approximation of the information in the database can lead to the collection of redundant segments in the database.
Nevertheless, the very efficient evaluation of the segment entropies outweighs the run-time penalty from including such redundant segments into the optimization

% Intuition experiment - entropy.
% \begin{figure}
% \centering
% \includegraphics[width=0.5\textwidth]{plots/entropy_increasing_excitation.eps}
% \caption{Experiment to illustrate the entropy as a segment selection score: The trajectory is divided into three parts: 1) movement only along global z-axis, approx. no rotation/translation on other axis, 2) only rotation around optical axis of the camera, 3) rotation and translation. It can be observed that the entropy reduces with increasing excitation; confirming the intuition that higher excitation yields more informative segments.}
% \label{fig:entropy_increasing}
% \end{figure}

%%%%%%%%%%%%%%%%%%%%%%%%%%%%%%%%%%%%%%%%%%%%%%%%%%%%%%%%%%%%%%%%%%%%%%%%%%%%%%%%%%%%%%%%%%%%%
\subsection{Sparsified Problem using Informative Segments}
\label{sec:sparified_problem}
\begin{figure}[]
\centering
\includegraphics[width=1.0\linewidth]{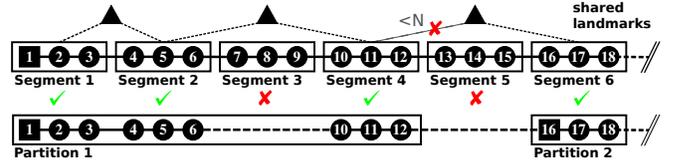}
\caption{The upper graph shows the full keyframe/landmark graph where the keyframes are assigned into fixed-size segments and the segments found to be informative are marked by green check-marks
The motion segments of the sparsified problem (shown below) can be partitioned into disjoint sets that are neither connected through inertial constraints nor share more than $N$ landmark observations with other partitions.
During optimization, the structurally unobservable states are fixed for exactly one keyframe per partition (marked by a square: e.g. 1)}.
\label{fig:segment_partitioning}
\end{figure}
%
% Problem definition: independend partitions.
%
The calibration over the set of informative segments differs from the full batch problem, described in \refsec{sec:batch_calib}, in that non-informative segments have been removed.
This results in missing inertial constraints between the remaining segments as shown in \reffig{fig:segment_partitioning} (e.g. between keyframe 6/10 and 12/16). 
The set of segments can then contain partitions that are neither constrained to other partitions through inertial constraints nor by joint landmark observations.
Each of these partitions can be seen as a (nearly) independent calibration problem only sharing calibration states with other partitions.

%
% Gauges of partitions.
%
If we assume the availability of sufficient landmark constraints and non-degenerate motion (e.g. only rotation), then the visual-inertial calibration problem contains two structurally unobservable states: the global orientation around the gravity vector and the global position.
To ensure an optimal and efficient convergence of the iterative solvers these redundant degrees of freedom need to be held constant during optimization for exactly one keyframe in each of the partitions.

\begin{algorithm}[]
 \footnotesize
 \KwIn{Set of motion segments $\mathcal{\mathbf{S}} = \{\mathcal{S}_0, ...,  \mathcal{S}_K\}$}
 \KwIn{Max. co-observed landmarks between partitions $N$}
 \KwResult{Set of motion segment partitions $\mathcal{P}$ }
 \SetKwFunction{Union}{Union}
 \SetKwFunction{Union}{MergePartitions}
 \SetKwFunction{Count}{CountSharedLandmarks}
 \BlankLine
 $\mathcal{\mathbf{P}} \leftarrow \{\}$\; \\
 \ForEach{$\mathcal{S}_k \in \mathcal{\mathbf{S}}$}  {
    $\mathcal{\mathbf{C}} \leftarrow \{\{\mathcal{S}_k\}\}$\; \\
    \ForEach{$p \in \mathcal{\mathbf{P}}$} {
        \If{\Count($p$, $\mathcal{S}_k$) $> N$} {
          $\mathcal{\mathbf{C}} \leftarrow \mathcal{\mathbf{C}} \cup \{p\}$ \;
		}
    }
   $p_{\mathcal{C}}$ $\leftarrow$ \Union{$\mathcal{\mathbf{C}}$}\; \\ 
   $\mathcal{\mathbf{P}} \leftarrow \left( \mathcal{\mathbf{P}} \setminus \mathcal{\mathbf{C}} \right) \cup \{p_{\mathcal{C}}\}$ \;
 }
 \caption{\small Partitioning segments on landmark co-visibility}
 \label{alg:union_find}
\end{algorithm}

% Union-find to find partitions.
Consequently, we identify these partitions by first joining all motion segments that have direct temporal neighbors into bigger segments (\reffig{fig:segment_partitioning}: e.g. segment 1 and 2).
At this point, all keyframes within the joined segments are constrained through inertial constraints.
The union-find algorithm, shown in \refalg{alg:union_find}, is then used to iteratively partition the segments into disjoint sets such that the count of co-observed landmarks between the partitions lies below a given threshold $N$ (here: 15).
This ensures that all keyframes within these partitions are either connected through inertial constraints or share sufficient landmark observations with other keyframes of the same partition.
Degenerate landmark configurations are theoretically possible, when using such a heuristic landmark threshold, but are highly unlikely and would only affect the convergence rate but not bias the estimates. 

% Bias constraints.
Additionally, a constraint between two bias states is introduced if keyframes were removed between the two (\reffig{fig:segment_partitioning}: e.g. between keyframe 6-10 and 12-16).
The bias evolution is modeled using a random walk as described in \refeq{sec:model_inertial}.

%% file: sections/4_results.tex
\begin{figure}[]
\centering
\includegraphics[width=0.9\linewidth]{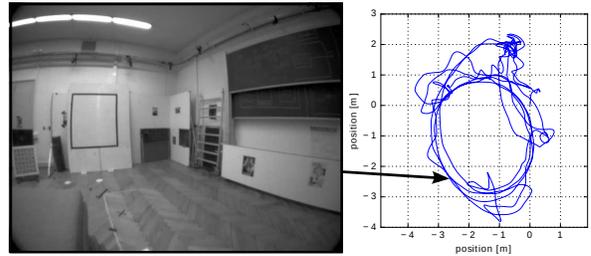}
\caption{Top-down view on the estimated trajectory of one of the evaluation datasets.}
\label{fig:experiment_room}
\end{figure}

% Experimental setup and Datasets.
A collection of $15$ datasets is used to assess the performance of the proposed visual-inertial self-calibration method.
The datasets were collected using a Google Tango Dev. Kit. tablet equipped with a MEMS IMU and a global shutter fisheye camera.
The device was hand-held while recording multiple trajectories of 3 min duration while freely moving in a room of approx. 8x6 m with a height of 4 m. 
The trajectories consist of calmer sections and sections that excite all rotational and translational degrees of freedom.
\reffig{fig:experiment_room} shows an image of the experimental environment together with a top-down view on one of the recorded trajectories.

In this section, we discuss our evaluation results based on these datasets along the following questions: 
\begin{itemize}
\item Does the sparsified calibration problem yield comparable results to the batch solution (\refsec{sec:performance_and_repeatabiliy})?
\item Is the proposed measure capable of identifying informative segments (\refsec{sec:least_info})?
\item Can the estimation be improved by grouping certain parameters and collecting segments for each group separately (\refsec{sec:eval_multiple_dbs})?
\item How does the proposed approach perform against comparable state-of-the-art methods in terms of run-time and estimation results (\refsec{sec:eval_rel_work})?
\end{itemize}

%%%%%%%%%%%%%%%%%%%%%%%%%%%%%%%%%%%%%%%%%%%%%%%%%%%%%%%%%%%%%%%%%%
\subsection{Performance and Repeatability of the Calibration}\label{sec:performance_and_repeatabiliy}
\begin{figure*}[]
\centering
\includegraphics[width=0.8\linewidth]{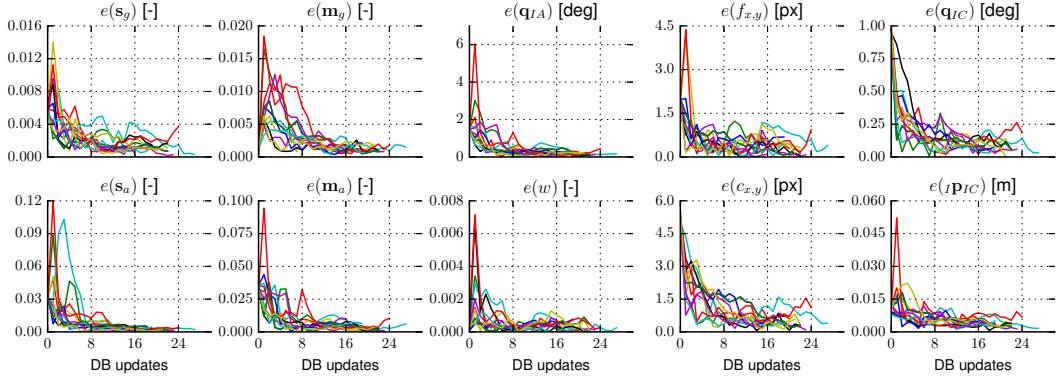}
\caption{Convergence of the calibration parameters for each update of the database i.e. when a new segment is added or a less informative segment is replaced. The y-axis shows the deviation to the batch solution as:  $e \left ( \mathbf{x} \right )  = \left \| \mathbf{\hat x}(k) \boxminus \mathbf{\hat x}_{batch} \right \|$.}
\label{fig:convergence}
\end{figure*}

% Statistics over N datasets.
We compare the estimated parameters of the sparsified problem to the batch solution that uses all keyframes.
The sparsified problem, here, denotes the calibration problem that only contains the most informative segments as described in \refsec{sec:method}.
The initial calibration states were set to the CAD values, if available, otherwise to the expected nominal values (i.e. no sensor misalignment, unit scale factors).
\reftab{tab:calib_states} shows the mean and standard deviation over the estimated parameters of the $15$ different datasets and the  convergence of the estimator is shown in \reffig{fig:convergence}.
The mean of rotation parameters corresponds to the Rodrigues angle $\gamma(\cdot)$ of the averaged quaternion \cite{markley2007averaging} over all data points and the standard deviation is calculated from the Rodrigues angles between the data points and the averaged quaternion.

These experiments show that the deviation between the sparsified estimation and the batch solution remains insignificant, in both the mean and standard deviation, even though large portions of the trajectory have been removed.
This indicates that the proposed method can sparsify the problem while retaining an estimation quality close the batch solution at a drastically reduced run-time.
It is important to note, that we cannot evaluate the accuracy of the estimated parameters as no ground-truth data is available, the statistics, however, give a good indication of the precision that can be achieved.

\input{tables/result-table.tex}

%%%%%%%%%%%%%%%%%%%%%%%%%%%%%%%%%%%%%%%%%%%%%%%%%%%%%%%%%%%%%%%%%%
\subsection{Evaluation of Segment Entropy to Select Informative Segments}\label{sec:least_info}
\begin{figure}[]
\centering
\includegraphics[width=1.0\linewidth]{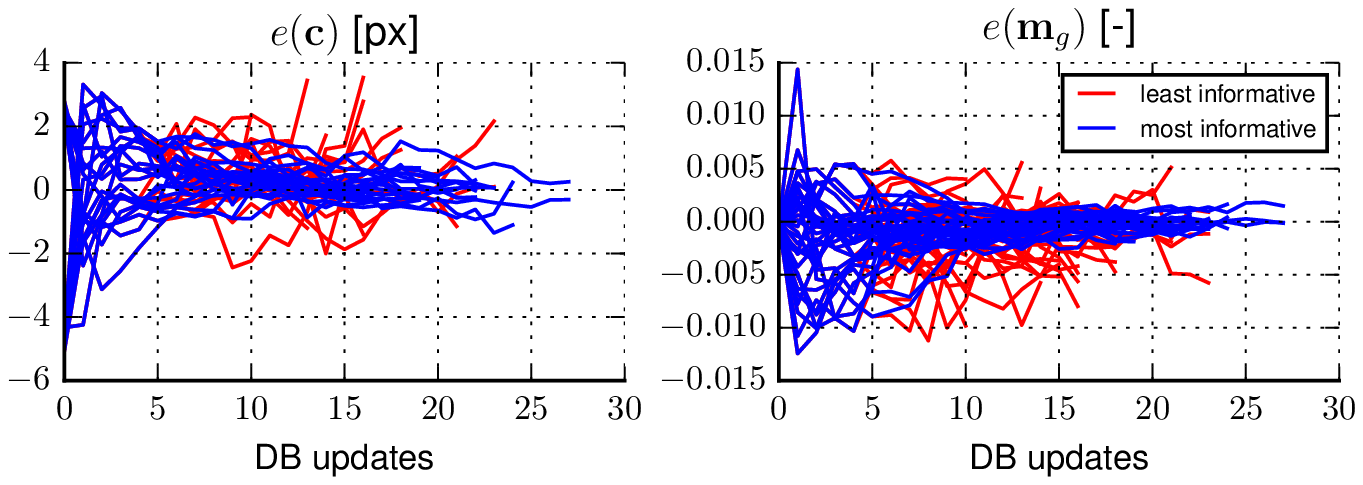}
%Autogenerated, do not change.
\resizebox{\columnwidth}{!}{%
	\begin{tabular}
	{ r r@{}c@{}l r@{}c@{}l r@{}c@{}l }\toprule
	&\multicolumn{3}{c}{\bf{most informative}} & \multicolumn{3}{c}{\bf{least informative}} & \multicolumn{3}{c}{\bf{batch}} \\ \midrule
	$\mathbf{c}$ [px] & 317.26 & ~$\pm$~ & 0.32 & 318.24 & ~$\pm$~ & 0.96 & 317.51 & ~$\pm$~ & 0.18 \\ 
	~ & 244.61 & ~$\pm$~ & 0.45 & 244.85 & ~$\pm$~ & 0.81 & 244.56 & ~$\pm$~ & 0.21 \\ 
	$\mathbf{m}_g$ [-] & 3.44e-04 & ~$\pm$~ & 6.48e-04 & -9.71e-04 & ~$\pm$~ & 2.60e-03 & 7.42e-05 & ~$\pm$~ & 3.23e-04 \\ 
	~ & 1.07e-03 & ~$\pm$~ & 8.49e-04 & 1.57e-04 & ~$\pm$~ & 2.59e-03 & 1.23e-03 & ~$\pm$~ & 4.75e-04 \\ 
	~ & 7.38e-04 & ~$\pm$~ & 8.36e-04 & -1.25e-03 & ~$\pm$~ & 3.75e-03 & 4.31e-04 & ~$\pm$~ & 5.02e-04 \\ 
	\bottomrule
	\end{tabular}
}%
\caption{Estimator performance when collecting the most-informative vs. the least-informative segments over $15$ datasets. The plots show the deviation $e(\cdot)$ of the proposed method to the batch solution.}
\label{fig:inv_entropy_conv}
\end{figure}

In this section, we conduct an experiment to investigate the suitability of the segment entropy to identify informative segments.
For this reason, we let the estimator from \refsec{sec:method} collect the least informative segments and compare the convergence of two parameters with results obtained by selecting the most informative segments.
In both cases, we collect 8 segments each of which consists of 40 keyframes resulting in a total of 320 keyframes which is equal to the data of 32 s.
The convergence is shown in \reffig{fig:inv_entropy_conv} together with the statistics on the final calibration states.
The calibration using the set of least-informative segments yields a higher estimation error \wrt the batch solution and a higher variance than the estimation using the most-informative segments.
This can be seen as an indication that:
\begin{enumerate}
\item the selection using the entropy identifies segments containing relevant information for sensor calibration,
\item the ratio of the number of selected segments to the total count of segments in the dataset is sufficiently low such that a careful selection is actually necessary.
\end{enumerate}

%%%%%%%%%%%%%%%%%%%%%%%%%%%%%%%%%%%%%%%%%%%%%%%%%%%%%%%%%%%%%%%%%%
\subsection{Influence of Multiple Parameter Groups}\label{sec:eval_multiple_dbs}
In this section, we analyze the effect on the estimation performance when collecting segments for individual parameter groups.
The estimator has been run with the parameter groups described in \refsec{sec:info_db} and as a comparison with a single group that contains all calibration parameters.

\begin{table}[]
\centering
\caption{Estimated calibration parameters: single parameter group vs. multiple groups.}
\label{tab:multi_queue}
	\resizebox{\columnwidth}{!}{%
		\begin{tabular}
		{ r  r@{}c@{}l r@{}c@{}l r@{}c@{}l}\toprule
		&\multicolumn{3}{c}{\bf{multiple groups}} & \multicolumn{3}{c}{\bf{single group}} & \multicolumn{3}{c}{\bf{batch}} \\ \midrule 
		$\mathbf{c}$ [px] & 317.35 & ~$\pm$~ & 0.21 & 317.31 & ~$\pm$~ & 0.35 & 317.51 & ~$\pm$~ & 0.18 \\ 
		~ & 244.61 & ~$\pm$~ & 0.29 & 244.51 & ~$\pm$~ & 0.52 & 244.56 & ~$\pm$~ & 0.21 \\ 
		$\mathbf{m}_g$ [-] & 1.64e-04 & ~$\pm$~ & 4.54e-04 & 2.60e-04 & ~$\pm$~ & 7.69e-04 & 7.42e-05 & ~$\pm$~ & 3.23e-04 \\ 
		~ & 1.09e-03 & ~$\pm$~ & 5.78e-04 & 1.05e-03 & ~$\pm$~ & 9.96e-04 & 1.23e-03 & ~$\pm$~ & 4.75e-04 \\ 
		~ & 5.33e-04 & ~$\pm$~ & 5.18e-04 & 8.68e-04 & ~$\pm$~ & 9.33e-04 & 4.31e-04 & ~$\pm$~ & 5.02e-04 \\ 
		\bottomrule
		\end{tabular}
	}%
\end{table}

The \reftab{tab:multi_queue} lists statistics of two estimated calibration parameters over $15$ datasets.
The results show that the variance of the estimates can be reduced by using multiple groups whereas the averaged error remains less affected.
Intuitively, this effect can be explained as follows:
If the problem structure contains groups of parameters that are rendered observable by different motion patterns and only a single informative database table is used then the chances are higher that only one type of motion is kept.
If multiple groups are used, however, the content in the database gets more stable and thus leads to a lower variance of the estimates.

%%%%%%%%%%%%%%%%%%%%%%%%%%%%%%%%%%%%%%%%%%%%%%%%%%%%%%%%%%%%%%%%%%
\subsection{Comparison to Related Method and Run-time}\label{sec:eval_rel_work}
%Experiment (describe jerome briefly, talk about differences)
In this section, we compare the proposed method against the work of \cite{maye2013self} in terms of estimation performance and run-time.
The latter work follows a similar approach that maintains a database of informative segments.
A calibration is run over the candidate segment and all segments already contained in the database.
The candidate is found to be informative if the information gain \wrt a calibration without the candidate segment lies above a certain threshold.
Since a complete calibration must be run for each candidate evaluation the complexity grows with each new segment in the database.
In contrast to the proposed method, this algorithm does consider all constraints when evaluating the information content of a candidate segment and does not make the assumption of segment independence as outlined in \refsec{sec:eval_entropy}.
Furthermore, they use a truncated QR instead of the Cholseky solver therefore it is more general and applicable for a wider range of problems although at a higher computational cost.

Two time points are given for the related work as it doesn't use an upper bound on the number of selected informative segments.
The first until the same amount of informative segments are collected as in the proposed method ($\approx$ 9) and the second (in brackets) until the information measure has been evaluated for each segment which is done in the proposed method by default.
The same $15$ datasets, used in the previous sections, have been processed with both methods.
\begin{table}[]
\centering
\caption{Run-time and number of processed segments for the three estimators. Each segment contains $40$ keyframes and corresponds to the data of $4$ s. Statistics are collected over $15$ datasets.}
\label{tab:run_time}
\begin{tabular}
{r  r@{}c@{}l r@{}c@{}l r@{}c@{}l}\toprule
&\multicolumn{3}{c}{\bf{our method}} & \multicolumn{3}{c}{\bf{related work} \cite{maye2013self}} & \multicolumn{3}{c}{\bf{batch}} \\ \midrule
\bf{num. segments} &  8.7 &~$\pm$~& 1.5 &    \multicolumn{3}{c}{9.0} &  38.0 &~$\pm$~& 3.2 \\ 
\bf{run-time [s]}  & 31.2  &~$\pm$~& 5.6  & 395.9  &~$\pm$~& 319.6  & 178.5 &~$\pm$~& 94.0 \\ 
&   & &   &  (7745.3   &~$\pm$~& 4601.7)  &  & &  \\ \bottomrule
\end{tabular}
\end{table}
The run-times are shown in \reftab{tab:run_time} and the estimated parameters in \reftab{tab:calib_states}.
The results show that the run-time of our algorithm is considerably lower than the full-batch and related work at very similar estimation performance.

%% file: tables/result-table.tex
\begin{table}[htpb]
	\scriptsize
	\centering
	\caption{Estimated calibration parameters for three different algorithms. The statistics are taken over $15$ datasets and show the mean and standard deviation. The number of used segments and run-time can be found in \reftab{tab:run_time}.}
	\label{tab:calib_states}
	\resizebox{\columnwidth}{!}{%
		\begin{tabular}
			{ r r@{}c@{}l r@{}c@{}l r@{}c@{}l }\toprule
			\shortstack{\bf{parameter}} &\multicolumn{3}{c }{\bf{informative}} & \multicolumn{3}{c }{\bf{all segments}} & \multicolumn{3}{c }{\bf{related work}} \\
			&\multicolumn{3}{c }{(proposed method)} & \multicolumn{3}{c }{(batch)} & \multicolumn{3}{c }{\cite{maye2013self}} \\
			\midrule
			$\mathbf{f}$ [px] & 254.71 & ~$\pm$~ & 0.28 & 254.50 & ~$\pm$~ & 0.13 & 254.89 & ~$\pm$~ & 0.35 \\ 
			~ & 254.63 & ~$\pm$~ & 0.29 & 254.47 & \!$\pm$\! & 0.14 & 254.68 & ~$\pm$~ & 0.32 \\ 
			$\mathbf{c}$ [px] & 317.26 & ~$\pm$~ & 0.32 & 317.51 & ~$\pm$~ & 0.18 & 317.74 & ~$\pm$~ & 0.35 \\ 
			~ & 244.61 & ~$\pm$~ & 0.45 & 244.56 & ~$\pm$~ & 0.21 & 242.87 & ~$\pm$~ & 0.67 \\ 
			$w$ [-] & 0.9222 & ~$\pm$~ & 0.0003 & 0.9222 & ~$\pm$~ & 0.0003 & 0.9227 & ~$\pm$~ & 0.0007 \\ 
			$\mathbf{s}_g - 1$ [-] & 6.17e-04 & ~$\pm$~ & 9.61e-04 & 4.45e-05 & ~$\pm$~ & 3.52e-04 & 5.40e-04 & ~$\pm$~ & 1.10e-03 \\ 
			~ & 5.80e-03 & ~$\pm$~ & 8.51e-04 & 5.56e-03 & ~$\pm$~ & 5.36e-04 & 4.78e-03 & ~$\pm$~ & 1.53e-03 \\ 
			~ & 8.54e-04 & ~$\pm$~ & 3.89e-04 & 8.44e-04 & ~$\pm$~ & 1.97e-04 & 4.00e-04 & ~$\pm$~ & 6.74e-04 \\ 
			$\mathbf{s}_a - 1$ [-] & -2.07e-02 & ~$\pm$~ & 2.28e-03 & -2.07e-02 & ~$\pm$~ & 1.47e-03 & -2.15e-02 & ~$\pm$~ & 2.33e-03 \\ 
			~ & -1.73e-02 & ~$\pm$~ & 1.25e-03 & -1.77e-02 & ~$\pm$~ & 5.21e-04 & -1.82e-02 & ~$\pm$~ & 1.01e-03 \\ 
			~ & -1.42e-02 & ~$\pm$~ & 1.34e-03 & -1.49e-02 & ~$\pm$~ & 5.39e-04 & -1.45e-02 & ~$\pm$~ & 1.08e-03 \\ 
			$\mathbf{m}_g$ [-] & 3.44e-04 & ~$\pm$~ & 6.48e-04 & 7.42e-05 & ~$\pm$~ & 3.23e-04 & 2.94e-04 & ~$\pm$~ & 7.52e-04 \\ 
			~ & 1.07e-03 & ~$\pm$~ & 8.49e-04 & 1.23e-03 & ~$\pm$~ & 4.75e-04 & 1.42e-03 & ~$\pm$~ & 1.16e-03 \\ 
			~ & 7.38e-04 & ~$\pm$~ & 8.36e-04 & 4.31e-04 & ~$\pm$~ & 5.02e-04 & 6.75e-04 & ~$\pm$~ & 5.51e-04 \\ 
			$\gamma(\mathbf{q}_{GA})$ [deg] & 1.467 & ~$\pm$~ & 0.141 & 1.498 & ~$\pm$~ & 0.056 & 1.501 & ~$\pm$~ & 0.060 \\ 
			$\mathbf{m}_a$ [-] & 1.78e-02 & ~$\pm$~ & 4.58e-03 & 1.79e-02 & ~$\pm$~ & 2.10e-03 & 1.82e-02 & ~$\pm$~ & 1.88e-03 \\ 
			~ & -2.91e-02 & ~$\pm$~ & 3.02e-03 & -2.95e-02 & ~$\pm$~ & 1.35e-03 & -3.00e-02 & ~$\pm$~ & 1.75e-03 \\ 
			~ & -1.18e-05 & ~$\pm$~ & 1.80e-03 & 1.13e-04 & ~$\pm$~ & 1.14e-03 & -1.62e-03 & ~$\pm$~ & 1.32e-03 \\ 
			$_{C}\mathbf{p}_{IC}$ [m] & 2.92e-03 & ~$\pm$~ & 2.79e-03 & 4.12e-03 & ~$\pm$~ & 1.01e-03 & -3.43e-03 & ~$\pm$~ & 3.42e-03 \\ 
			~ & 1.25e-02 & ~$\pm$~ & 2.55e-03 & 1.34e-02 & ~$\pm$~ & 1.37e-03 & 1.38e-02 & ~$\pm$~ & 1.94e-03 \\ 
			~ & -5.47e-03 & ~$\pm$~ & 2.86e-03 & -5.68e-03 & ~$\pm$~ & 1.14e-03 & -2.81e-03 & ~$\pm$~ & 3.01e-03 \\ 
			$\gamma(\mathbf{q}_{IC})$ [deg] & 0.311 & ~$\pm$~ & 0.062 & 0.306 & ~$\pm$~ & 0.019 & 0.170 & ~$\pm$~ & 0.047 \\ 
			\bottomrule
		\end{tabular}
	}%
\end{table}

%% file: sections/5_conclusions.tex
%
% Summary
%
In this work, we presented a novel method for efficient self-calibration of visual-inertial sensor systems that runs in parallel to an existing SLAM system.
An information-theoretic measure is introduced to evaluate the information content of motion segments keeping a fixed number of the most-informative segments in a database.
The proposed measure can be efficiently evaluated without running an expensive batch calibration beforehand.
Once the database contains enough data, an optimization is run over these segments to update the calibration parameters.

Real-world experiments show that the sparsified problem yields similar results to the full batch solution at a significantly reduced computational cost.
Even, when compared to previous work on segment based calibration, our approach shows a reduction of the run-time by a factor of approx. $10$.
Therefore, the proposed method is well suited for performing self-calibration on resource constrained platforms and can enable accurate operation over the entire lifespan.

%% file: sections/6_acknowledgement.tex
The research leading to these results has received funding from Google's project Tango.

%% file: ms.bbl
\begin{thebibliography}{10}

\bibitem{li2014high}
M.~Li, H.~Yu, X.~Zheng, and A.~I. Mourikis, ``High-fidelity sensor modeling and
  self-calibration in vision-aided inertial navigation,'' in {\em ICRA},
  pp.~409--416, IEEE, 2014.

\bibitem{leutenegger2015keyframe}
S.~Leutenegger, S.~Lynen, M.~Bosse, R.~Siegwart, and P.~Furgale,
  ``Keyframe-based visual--inertial odometry using nonlinear optimization,''
  {\em IJRR}, vol.~34, no.~3, pp.~314--334, 2015.

\bibitem{li2013high}
M.~Li and A.~I. Mourikis, ``High-precision, consistent {EKF}-based
  visual--inertial odometry,'' {\em The International Journal of Robotics
  Research}, vol.~32, no.~6, pp.~690--711, 2013.

\bibitem{bloesch2015robust}
M.~Bloesch, S.~Omari, M.~Hutter, and R.~Siegwart, ``Robust visual inertial
  odometry using a direct {EKF}-based approach,'' in {\em IROS}, pp.~298--304,
  IEEE, 2015.

\bibitem{zhang2000flexible}
Z.~Zhang, ``A flexible new technique for camera calibration,'' {\em TPAMI},
  vol.~22, no.~11, pp.~1330--1334, 2000.

\bibitem{devernay2001straight}
F.~Devernay and O.~Faugeras, ``Straight lines have to be straight,'' {\em
  Machine vision and applications}, vol.~13, no.~1, pp.~14--24, 2001.

\bibitem{krebs}
C.~Krebs and J.~Rehder, ``Generic imu-camera calibration algorithm,'' {\em
  Semester thesis}, 2012.

\bibitem{nikolic2016non}
J.~Nikolic, M.~Burri, I.~Gilitschenski, J.~Nieto, and R.~Siegwart,
  ``Non-parametric extrinsic and intrinsic calibration of visual-inertial
  sensor systems,'' {\em Sensors Journal}, vol.~16, no.~13, pp.~5433--5443,
  2016.

\bibitem{rehder2016extending}
J.~Rehder, J.~Nikolic, T.~Schneider, T.~Hinzmann, and R.~Siegwart, ``Extending
  kalibr: Calibrating the extrinsics of multiple imus and of individual axes,''
  in {\em ICRA}, IEEE, 2016.

\bibitem{mirzaei2008kalman}
F.~M. Mirzaei and S.~I. Roumeliotis, ``A {K}alman filter-based algorithm for
  imu-camera calibration: Observability analysis and performance evaluation,''
  {\em IEEE TOR}, vol.~24, no.~5, pp.~1143--1156, 2008.

\bibitem{maye2013self}
J.~Maye, P.~Furgale, and R.~Siegwart, ``Self-supervised calibration for robotic
  systems,'' in {\em Intelligent Vehicles Symposium (IV)}, pp.~473--480, IEEE,
  2013.

\bibitem{keivan2014constant}
N.~Keivan and G.~Sibley, ``Constant-time monocular self-calibration,'' in {\em
  ROBIO)}, pp.~1590--1595, IEEE, 2014.

\bibitem{hausman2016observability}
K.~Hausman, J.~Preiss, G.~Sukhatme, and S.~Weiss, ``Observability-aware
  trajectory optimization for self-calibration with application to uavs,'' {\em
  arxiv}, 2016.

\bibitem{trawny2005indirect}
N.~Trawny and S.~I. Roumeliotis, ``Indirect kalman filter for 3d attitude
  estimation,'' {\em University of Minnesota, Dept. of Comp. Sci. \& Eng.,
  Tech. Rep}, vol.~2, 2005.

\bibitem{ceres_solver}
S.~Agarwal, K.~Mierle, and Others, ``Ceres solver.''
  \url{http://ceres-solver.org}.

\bibitem{nerurkar2014c}
E.~D. Nerurkar, K.~J. Wu, and S.~I. Roumeliotis, ``C-klam: Constrained
  keyframe-based localization and mapping,'' in {\em ICRA}, pp.~3638--3643,
  IEEE, 2014.

\bibitem{lynen2015get}
S.~Lynen, T.~Sattler, M.~Bosse, J.~Hesch, M.~Pollefeys, and R.~Siegwart, ``Get
  out of my lab: Large-scale, real-time visual-inertial localization,'' in {\em
  RSS}, 2015.

\bibitem{hughes2010equidistant}
C.~Hughes, P.~Denny, M.~Glavin, and E.~Jones, ``Equidistant fish-eye
  calibration and rectification by vanishing point extraction,'' {\em TPAMI},
  vol.~32, no.~12, pp.~2289--2296, 2010.

\bibitem{markley2007averaging}
F.~L. Markley, Y.~Cheng, J.~L. Crassidis, and Y.~Oshman, ``Averaging
  quaternions,'' {\em Journal of Guidance, Control, and Dynamics}, vol.~30,
  no.~4, pp.~1193--1197, 2007.

\end{thebibliography}
